\documentclass[10pt]{article} 
\usepackage{changepage} 
\usepackage{float}
\usepackage[preprint]{tmlr}
\pdfoutput=1
\usepackage{caption}

\usepackage{pgfplots}
\pgfplotsset{compat=1.17} 

\usepackage{graphicx}
\usepackage{ifthen}

\usepackage{multirow}
\usepackage{booktabs}

\newboolean{isFinal}
\setboolean{isFinal}{false}

\usepackage{listings}

\usepackage{enumitem}
\usepackage{algorithm}
\usepackage{algpseudocode}

\usepackage{amsmath,amsfonts,bm}

\def\eqref#1{equation~\ref{#1}}

\def\1{\bm{1}}

\DeclareMathAlphabet{\mathsfit}{\encodingdefault}{\sfdefault}{m}{sl}
\SetMathAlphabet{\mathsfit}{bold}{\encodingdefault}{\sfdefault}{bx}{n}

\usepackage{url}

\usepackage{hyperref}

\lstdefinestyle{mystyle}{
    basicstyle=\ttfamily\footnotesize,   
    numbers=left,                        
    numberstyle=\tiny,                   
    stepnumber=1,                         
    numbersep=5pt,                        
    backgroundcolor=\color{gray!10},      
    keywordstyle=\color{blue}\bfseries,   
    commentstyle=\color{green!50!black},  
    stringstyle=\color{red},              
    breaklines=true,                      
    tabsize=4,                            
    frame=single                          
}

\usepackage{natbib}

\title{CART-ELC: Oblique Decision Tree Induction via Exhaustive Search}

\author{\name Andrew D. Laack \email andrew@laack.co\\
      University of Wisconsin-Superior
}

\def\month{MM}  
\def\year{YYYY} 
\def\openreview{\url{https://openreview.net/forum?id=XXXX}} 

\begin{document}

\ifthenelse{\boolean{isFinal}}{

	{\large \textbf{CART-ELC: Oblique Decision Tree Induction via Exhaustive Search}} \\
	{ {Andrew D. Laack}}

    \section{Prerequisites}

	Terms a broad computer science audience may not be familiar with are defined in this section.

    \begin{enumerate}
        \item \textbf{Machine Learning:} 
			A field that focuses on algorithms that learn from patterns in data.
        \item \textbf{Sample:} 
			A single data point, often the result of an experiment. Mathematically, samples are vectors. 
        \item \textbf{Feature:}  
			A coordinate of a sample, or an attribute of all samples.
		\item \textbf{Label:}  
			Label and classification are used interchangeably to describe the value we are trying to predict about a sample.
        \item \textbf{Model:}  
			A model is a trained implementation of a machine learning algorithm that has learned patterns from data.
        \item \textbf{Decision Tree:}  
			A machine learning algorithm that uses a binary tree for classification (discrete labels) or regression (continuous labels) where tree traversals are done based on feature values for the sample being evaluated. Additionally, decision trees generally use axis-aligned splitting boundaries. Pseudocode for a decision tree algorithm is found in \autoref{decision_tree}.
		\item \textbf{Splitting Criterion:}  
			A function to quantify how good a candidate split is for a decision tree (see \autoref{splitting_criteria}).
        \item \textbf{Hyperplane:}  
			A mathematical object that has $n-1$ dimensions in $n$ dimensional space.
        \item \textbf{Oblique Decision Tree:}  
			A decision tree that uses hyperplanes as splitting boundaries that are not constrained to axis-alignment.
        \item \textbf{Accuracy:}  
			The relative frequency of correct predictions made by a model on a given dataset.
        \item \textbf{P-value:}  
			The probability of obtaining a result at least as strong as the one observed, assuming the null hypothesis is true.
        \item \textbf{Cohen's d:}  
			A measure of effect size quantifying the difference in standard deviations between two means.
    \end{enumerate}

	\newpage

\subsection{Decision Tree Pseudocode}\label{decision_tree}

\begin{algorithm}
	\caption{CART Algorithm}
\begin{algorithmic}[1]
\Function{fit}{samples, labels, featureCount}
    \If{homogeneous(labels)}
        \State \Return Node(majorityClass(labels)) 
    \EndIf
	\State bestSplit, bestSplittingScore $\gets$ None, worstSplittingScore()
    \For{sample in samples}
		\For{feature in range(0,featureCount)}
			\State currentSplit $\gets$ (feature, sample[feature])
            \State currentSplittingScore $\gets$ evaluateSplit(currentSplit, samples)
			\If{ isBetterThan(currentSplittingScore, bestSplittingScore)}
                \State bestSplittingScore, bestSplit  $\gets$ currentSplittingScore, currentSplit
            \EndIf
        \EndFor
    \EndFor
    \State left, right $\gets$ splitDataByBestSplit(samples, labels, bestSplit)
	\If{left is empty or right is empty}
		\State \Return Node(majorityClass(labels)) 
	\EndIf
    \State leftSubtree $\gets$ fit(left.samples, left.labels, featureCount)
    \State rightSubtree $\gets$ fit(right.samples, right.labels, featureCount)
    \State tree $\gets$ Node(bestSplit)
    \State tree.left, tree.right $\gets$ leftSubtree, rightSubtree
    \State \Return tree
\EndFunction
\end{algorithmic}
\end{algorithm}

\vfill

    \begin{center}
		\textbf{\large This concludes the Prerequisites section.}
    \end{center}

\newpage
}{}

\maketitle


\begin{abstract}

Oblique decision trees have attracted attention due to their potential for improved classification performance over traditional axis-aligned decision trees. However, methods that rely on exhaustive search to find oblique splits face computational challenges. As a result, they have not been widely explored. We introduce a novel algorithm, Classification and Regression Tree - Exhaustive Linear Combinations (CART-ELC), for inducing oblique decision trees that performs an exhaustive search on a restricted set of hyperplanes. We then investigate the algorithm's  computational complexity and its predictive capabilities. Our results demonstrate that CART-ELC consistently achieves competitive performance on small datasets, often yielding statistically significant improvements in classification accuracy relative to existing decision tree induction algorithms, while frequently producing shallower, simpler, and thus more interpretable trees.

\end{abstract}

%
%

\section{Introduction}\label{introduction}

Decision tree induction algorithms are among the most widely used algorithms for training interpretable machine learning models. This interpretability is one of the reasons they are often preferred over other models with better predictive capabilities \citep{survey, boosting}. While there has been much research on axis-aligned decision trees \citep{recent_overview, breiman1984cart, survey}, research on oblique decision trees has been less extensive \citep{HHCart, MurthyKS94, breiman1984cart}.

Where traditional decision trees use axis-aligned splits to partition the feature space, oblique decision trees use linear combinations of features that are not restricted to axis-alignment. This allows oblique decision trees to be stronger learners with shallower trees while still being fairly interpretable \citep{MurthyKS94}. However, this increased flexibility results in a larger search space as the number of hyperplanes that uniquely partition a dataset when considering the relative positions of samples with respect to the hyperplane is at most $2^m \cdot \binom{n}{m}$,  where $m$ is the dimensionality of the feature space and $n$ is the number of samples \citep{MurthyKS94}.

\subsection{Related Works}\label{related_works}

  \subsubsection{CART}

The Classification and Regression Tree (CART) algorithm \citep{breiman1984cart} induces axis-aligned decision trees by recursively partitioning the dataset. At each step, it evaluates a set of candidate axis-aligned splits, selects the best one, and partitions the dataset using this split. These steps are then performed recursively for each partition until a stopping criterion is met.

  \subsubsection{CART-LC}

The Classification and Regression Tree - Linear Combinations (CART-LC) algorithm \citep{breiman1984cart} is a deterministic hill climbing algorithm for inducing oblique decision trees. The algorithm starts by median-centering the features and dividing them by their interquartile range. It then identifies the best axis-aligned split and iteratively adjusts the coefficients of the corresponding hyperplane to improve the split. Once improvements fall below a threshold, the algorithm converts the refined split from the transformed feature space to the original feature space, and uses this split to partition the dataset. This process is then applied recursively for each partition until a stopping criterion is met.

  \subsubsection{OC1}

The Oblique Classifier 1 (OC1) algorithm \citep{MurthyKS94} is a stochastic hill climbing algorithm for inducing oblique decision trees. It begins by initializing a random hyperplane and deterministically perturbs the hyperplane's coefficients to improve its splitting score. If no improvements can be made, a new hyperplane is created by adding random values to the coefficients in an attempt to escape the local optimum. The new hyperplane is then evaluated against the previous one. If the new score is better, the process repeats, starting with this new hyperplane as the initial hyperplane. If it is not better, another random hyperplane is initialized, and the entire procedure starts again. This process of applying perturbations and then attempting to escape local optima is applied $z$ times, where $z$ is a hyperparameter of the algorithm. After $z$ iterations, the split with the best splitting score is then used to partition the dataset. The algorithm then recursively applies the above procedure to each partition until a stopping criterion is met.

\subsubsection{HHCART}

The Householder Classification and Regression Tree (HHCART) algorithm \citep{HHCart} induces oblique decision trees by leveraging class-specific orientation. HHCART(D), a variant of the algorithm, computes the dominant eigenvector of the covariance matrix for each class. Each of these dominant eigenvectors is then individually used to perform a Householder transformation to align the corresponding class orientation with a coordinate axis. In the resulting feature space, the algorithm searches for the best axis-aligned split, which is then mapped back to the original feature space as an oblique split. The best oblique split found after evaluating each dominant eigenvector is used to partition the dataset. This process is then applied to each of the resulting partitions until a stopping criterion is met. \citet{HHCart} defined two variants of HHCART: HHCART(D), which only uses the dominant eigenvector for each class, and HHCART(A), which uses all eigenvectors for each class.

\subsection{Limitations}\label{Limitations}

A limitation of the CART algorithm is that it only considers axis-aligned splits. These splits can be suboptimal when the data is linearly separable, as described by \citet{breiman1984cart}. Algorithms like CART-LC and OC1 are not constrained to axis-aligned splits, but they often converge to local optima due to their hill climbing nature. While OC1 attempts to escape local optima by adding random values to the coefficients of the splitting hyperplane after its improvements have stagnated, this strategy does not guarantee escaping local optima. HHCART is not limited in either of these regards, but it does only test splits that are axis-aligned in specific transformed feature spaces. This subset of oblique splits can be overly limited in cases where it is computationally feasible to evaluate more oblique splits.

Additionally, while CART-LC, OC1, and HHCART can generate shallower trees relative to CART \citep{MurthyKS94}, the resulting trees often use splits that are linear combinations of many features. This results in oblique trees that can be difficult to interpret, and more costly to perform predictions with.

\section{CART-ELC}\label{algorithm}

\subsection{Description}\label{Description}

We introduce CART-ELC to address the limitations described in \autoref{Limitations}. CART-ELC has one hyperparameter, $r$, defining both the minimum number of samples each candidate hyperplane must pass through and the maximum number of non-zero coefficients the hyperplane's normal vector can have. We define $m$ as the dimensionality of the feature space. With these definitions, we now discuss the two cases for the algorithm.

\textbf{Case 1: $\mathbf{r = m}$}

\begin{adjustwidth}{1em}{1em}

In the general case for $r=m$, CART-ELC performs an exhaustive search of all unique hyperplanes that pass through at least $r$ samples. This is done by evaluating each hyperplane according to a splitting criterion, tracking the best splitting hyperplane, and after all hyperplanes have been evaluated, the algorithm partitions the dataset using the best split. This process is then recursively applied to both partitions, stopping when a stopping criterion is met. The general case for $r=m$ is when each selection of $r$ samples uniquely identifies a hyperplane. In such cases, this process guarantees the greedily optimal hyperplane passing through at least $r$ samples will be used to partition the dataset at each splitting node. This guarantee does not apply in degenerate cases where samples don't uniquely identify a hyperplane, but when floating-point vectors are sampled from a continuous distribution, it is unlikely a selection of $r$ samples is affinely dependent. In such cases, the normal vector for a hyperplane passing through the $r$ samples is chosen according to the deterministic tie-breaking behavior of the underlying linear algebra solver's implementation. 


\end{adjustwidth}

\textbf{Case 2: $\mathbf{r < m}$}

\begin{adjustwidth}{1em}{1em}

	CART-ELC limits the search space to hyperplanes with at most $r$ non-zero coefficients and one bias term. In the general case, the algorithm considers each unique hyperplane defined by selecting a combination of $r$ features passing through at least $r$ samples where the remaining $m-r$ features have coefficients of zero. Like the $r=m$ case, each candidate is then evaluated using a splitting criterion with the best split chosen to partition the dataset. This process is then recursively applied to both partitions until a stopping criterion is met. Additionally, like the $r=m$ case, the deterministic tie-breaking behavior of the underlying linear algebra solver's implementation is used to select a normal vector for a hyperplane that passes through the $r$ samples when the selection of samples and features is affinely dependent.

\end{adjustwidth}

\textbf{Hyperplane Orientation}

\begin{adjustwidth}{1em}{1em}

	Every hyperplane has two unit normal vectors in $\mathbb{R}^n$ for $n \in \mathbb{N}$.  Since we assign samples to the left partition when samples are on or above the hyperplane, the selection of a normal vector can change how samples that lie on the hyperplane contribute to the splitting score. To eliminate this nondeterminism, one may enforce a canonical sign rule. 
Given this, optimality guarantees for hyperplane splitting only hold under the deterministic orientation of the hyperplanes.

\end{adjustwidth}

\begin{figure}[h]
    \centering
        \centering
        \begin{tikzpicture}[scale=1]
            \draw[->] (-1,0) -- (4,0) node[right] {$x$};
            \draw[->] (0,-1) -- (0,4) node[above] {$y$};

            \fill[black] (1,1) circle (2pt) node[above] {$x_1$};
            \fill[black] (4,4) circle (2pt) node[below] {$x_2$};

            \draw[thick, dashed] (0,1.25) -- (4,3.25) node[right] {$H_1$};
            \draw[thick, dashed] (0,2.5) -- (4,4.5) node[right] {$H_2$};
        \end{tikzpicture}
        \caption{Hyperplanes resulting in unique partitionings of two samples.}
		\label{fig:hyper}
\end{figure}
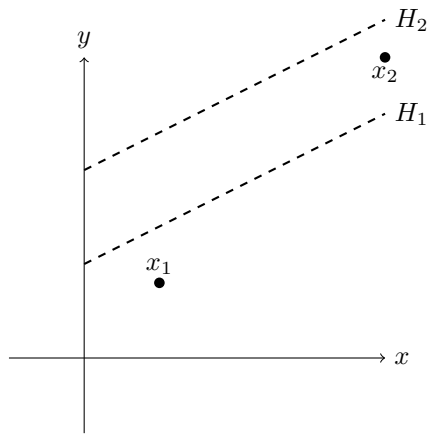

In the general case for cases 1 and 2, the algorithm guarantees finding the best splitting hyperplane with at most $r$ non-zero coefficients and a bias term that passes through at least $r$ samples at each splitting node, but in neither case can it guarantee this is the greedily optimal oblique split. An illustration of this is shown in \autoref{fig:hyper}. Based on the samples in \autoref{fig:hyper}, if we define $r = m$, only one hyperplane will be evaluated: the hyperplane that passes through both samples and places them in the same partition. This hyperplane will result in the same partitioning as the hyperplane $H_2$. However, a hyperplane that results in the same partitioning as hyperplane $H_1$ is better if $x_1$ and $x_2$ have different labels. Since CART-ELC does not evaluate a hyperplane that partitions the samples in this way, the algorithm can not guarantee greedily optimal oblique splits. 

Additionally, there may be multiple selections of features and samples that describe the same hyperplane. While computing the splitting score for each hyperplane is not necessary, the algorithm will evaluate each. When floating-point vectors are sampled from a continuous distribution, it is infrequent different selections of samples and features will result in the same hyperplane.

The algorithm also defines two stopping criteria. The first stopping criterion is class homogeneity: if a partition contains samples that all belong to the same class, the algorithm will stop because subsequent splits would not change the model's predictions.

The second criterion is when a partition is empty. If no special handling is implemented, the algorithm will attempt to partition samples indefinitely in cases where one partition is empty and the other is not homogeneous. As such, the algorithm must verify both partitions are not empty after each partitioning.

Finally, the choice to not specify a splitting criterion, additional stopping criteria, specific hyperplane tie-breaking behavior, or pruning was made to ensure future research can use this algorithm without limiting its usefulness to specific domains.

\subsection{Pseudocode}\label{Pseudocode}

\begin{algorithm}[H]
	\caption{CART-ELC}
\begin{algorithmic}[1]
\Function{fit}{samples, labels, $m$, $r$}
    \If{homogeneous(labels)}
        \State \Return Node(majorityClass(labels)) 
    \EndIf
	\State bestSplit, bestSplittingScore $\gets$ None, worstSplittingScore()
    \For{selectedSamples in combinations(samples, $r$)}
        \For{selectedFeatures in combinations($m$, $r$)}
            \State vectorsToPassThrough $\gets$ featureSubset(selectedSamples, selectedFeatures)
            \State currentSplit $\gets$ findHyperplanePassingThrough(vectorsToPassThrough)
            \State currentSplittingScore $\gets$ evaluateSplit(currentSplit, samples)
			\If{ isBetterThan(currentSplittingScore, bestSplittingScore)}
                \State bestSplittingScore, bestSplit  $\gets$ currentSplittingScore, currentSplit
            \EndIf
        \EndFor
    \EndFor
    \State left, right $\gets$ splitDataByBestSplit(samples, labels, bestSplit)
	\If{left is empty or right is empty}
        \State \Return Node(majorityClass(labels)) 
	\EndIf
    \State leftSubtree $\gets$ fit(left.samples, left.labels, $m$, $r$)
	\State rightSubtree $\gets$ fit(right.samples, right.labels, $m$, $r$)
    \State tree $\gets$ Node(bestSplit)
    \State tree.left, tree.right $\gets$ leftSubtree, rightSubtree
    \State \Return tree
\EndFunction
\end{algorithmic}
\end{algorithm}

\subsection{Splitting Criteria}\label{splitting_criteria}

To evaluate the time complexity of our algorithm and perform an empirical analysis, we must define some splitting criteria. The three we implemented and will define are the twoing criterion \citep{breiman1984cart}, the Gini criterion \citep{breiman1984cart}, and information gain \citep{quinlan1986}. The default splitting criterion for our implementation of CART-ELC is the Gini criterion. This decision is based on our empirical analysis in \autoref{empirical_criteria_comparison}, which showed the twoing and Gini criteria often result in trees that yield the same classification accuracy, but the Gini criterion is computationally cheaper. Despite this, our empirical evaluation in \autoref{empirical_comparison} uses the twoing criterion to maintain consistency with prior research.

\subsubsection{Twoing Criterion}\label{twoing}

The twoing criterion is a splitting criterion that evaluates the quality of a split by considering the number of samples on each side of the splitting hyperplane and the difference in the distribution of classifications between the two partitions. Trees using the twoing criterion tend to be well-balanced because the splitting score is maximized when the number of samples in each partition is equal.

The formula for the twoing criterion defines the following variables:

\begin{itemize}
    \setlength{\itemsep}{0.1em} 
    \item \( p_L \): The probability a sample is in the left partition.
    \item \( p_R \): The probability a sample is in the right partition.
    \item \( C \): The set of all classes in the union of the left and right partitions.
    \item \( p_{Lj} \): The probability a random sample from the left partition has a classification of \( j \).
    \item \( p_{Rj} \): The probability a random sample from the right partition has a classification of \( j \).
\end{itemize}

\begin{equation*}
T = \frac{p_L p_R}{4} \left( \sum_{j \in C} \left| p_{Lj} - p_{Rj} \right| \right)^2
	\label{twoing_equ}
\end{equation*}

\subsubsection{Gini Criterion}\label{gini}

The Gini criterion is the most popular splitting criterion for decision trees. Trees induced using the Gini criterion minimize class impurity, resulting in trees that may be less balanced than the twoing criterion. The splitting score is calculated by computing the Gini impurity for both partitions, weighting them based on the number of samples in each partition, and summing the results.

The formula for the Gini criterion defines the following variables:

\begin{itemize}
    \setlength{\itemsep}{0.1em} 
    \item \( p_L \): The probability a sample is in the left partition.
    \item \( p_R \): The probability a sample is in the right partition.
    \item \( C \): The set of all classes in the union of the left and right partitions.
    \item \( p_{Lj} \): The probability a random sample from the left partition has a classification of \( j \).
    \item \( p_{Rj} \): The probability a random sample from the right partition has a classification of \( j \).
\end{itemize}

\begin{equation*}
	G = p_L \left(1 - \sum_{j\in C} p_{Lj}^2 \right) + p_R \left(1 - \sum_{j \in C} p_{Rj}^2 \right)
	\label{gini_equ}
\end{equation*}

\subsubsection{Information Gain}\label{IG}

Information gain is a criterion based on reducing class impurity. Unlike the Gini criterion, information gain calculates impurity using entropy.

The formula for information gain defines the following variables where $H(S)$ is the entropy of the set $S$ \citep{it}:

\begin{itemize}
    \setlength{\itemsep}{0.1em} 
    \item \(S \): The set of all samples before the split.
    \item \(S_L \): The set of all samples in the left partition.
    \item \(S_R \): The set of all samples in the right partition.
\end{itemize}

\begin{equation*}
	IG = H(S) - \left( \frac{|S_{L}|}{|S|} H(S_{L}) + \frac{|S_{R}|}{|S|} H(S_{R}) \right)
\end{equation*}

\subsection{Time Complexity}\label{complexity}

\subsubsection{Asymptotic Time Complexity Analysis}\label{analysis}

To evaluate the algorithm's asymptotic time complexity, we concentrate on the cost of finding and executing a split at a single decision node. Although a complete decision tree often contains multiple splitting nodes, the computational cost is most significant for the first split. Once the dataset is partitioned, subsequent splits operate on smaller subsets and thus require less computation. This is shown in \autoref{fig:operations} by considering the computational differences between the $n=1000$ and $n=500$ columns.

The homogeneity calculation, which requires at most \( n \) operations per label, is dominated by the complexity of the loop over each hyperplane. As such, it does not need to be considered for the final asymptotic time complexity. Next, we consider the evaluation of hyperplanes. Assume the feature matrix is \( n \times m \), and 

\begin{equation*}
	0 < r \leq n, m. 
\end{equation*}

Hyperplane construction requires selecting \( r \) samples from \( n \), which can be done in \( \binom{n}{r} \) ways, and for each selection of samples, there are \( \binom{m}{r} \) ways to choose \( r \) features. Therefore, the total number of hyperplanes to evaluate is

\[
	\binom{n}{r} \cdot\binom{m}{r}.
\]

To derive the equation of the hyperplane passing through the \(r\) samples, an \( r \times r \) matrix is constructed based on the $r$ selected features and samples. Subsequently, its column-wise means are computed; these means are then subtracted from each row of the matrix, and the covariance matrix is calculated. Eigen decomposition is then performed on the resulting matrix to find the hyperplane's normal vector, and the hyperplane's bias term is determined by taking the dot product between the normal vector and the column-wise mean vector. These calculations are sufficient to derive the hyperplane's equation in normal form. The calculation of the covariance matrix and the eigen decomposition dominate the complexity of finding the equation for the hyperplane. Each is in the complexity class of \( \Theta(r^3) \), assuming the use of a naïve matrix multiplication approach \citep{matrix}.

We then evaluate the split using our splitting criterion. Assume our splitting criterion is the Gini criterion. To compute our splitting score, we evaluate

\begin{equation*}
	G = p_L \left(1 - \sum_{j\in C} p_{Lj}^2 \right) + p_R \left(1 - \sum_{j \in C} p_{Rj}^2 \right).
	\label{gini_usage}
\end{equation*}

The complexity of calculating the splitting score is dominated by determining which side of the hyperplane each sample lies on. This requires calculating the dot product between the sample and the hyperplane's normal vector, adding the bias term, and checking if the result is positive, negative, or zero. As such, the evaluation of one hyperplane is in the complexity class of \( \Theta(rn) \), and since \( |C| \leq n \), the number of classes does not impact the asymptotic time complexity of evaluating each hyperplane.

To split the dataset on the best split, we recompute the side of the hyperplane each sample lies on and add each sample to the corresponding list for samples that are either above or on the hyperplane or strictly below it. While this recomputation can be avoided by caching the results, it does not impact the overall asymptotic time complexity of the algorithm. The evaluation of all samples will take \( \Theta(rn) \) time, and since we need to move each sample to the appropriate list, and each sample has \( m \) features, this will take \( \Theta(nm) \) time. Thus, the asymptotic time complexity of splitting the dataset is in the complexity class of

\begin{equation*}
	\Theta(rn + nm).
\end{equation*}

Since splitting on the best split is only done once, its complexity is dominated by the complexity of evaluating all hyperplanes. As such, this final split does not impact the overall asymptotic time complexity.

Since there are \( \binom{n}{r} \cdot \binom{m}{r} \) splits to evaluate, each split takes \( \Theta(r^3) \) operations to compute the hyperplane, and \( \Theta(rn) \) time to evaluate, we find the asymptotic time complexity of finding and executing a split at a single decision node is in the complexity class of

\[
	\Theta\left( \binom{n}{r} \cdot \binom{m}{r} \cdot r(r^2 + n) \right).
\]

\subsubsection{Operations for a Single Split}\label{ops}

\begin{table}[h]
	\centering
    \caption{Operations required for CART-ELC to identify and execute a split at a single decision node, assuming a multiplicative factor of one for the asymptotic time complexity, no additive constants, and $r = m$.}
	\small
	\begin{tabular}{lcccccc} 
		\addlinespace
		\toprule
		\multirow{2}{*}{r} & \multicolumn{6}{c}{n}  \\ 
		\cmidrule(lr){2-7}  
		& 100 & 500 & 1000 & 5000 & 10000 & 20000 \\ 
		\midrule
			1 & 1.01e+04 & 2.50e+05 & 1.00e+06 & 2.50e+07 & 1.00e+08 & 4.00e+08 \\
			2 & 1.03e+06 & 1.26e+08 & 1.00e+09 & 1.25e+11 & 1.00e+12 & 8.00e+12 \\
			3 & 5.29e+07 & 3.16e+10 & 5.03e+11 & 3.13e+14 & 5.00e+15 & 8.00e+16 \\
			4 & 1.82e+09 & 5.31e+12 & 1.68e+14 & 5.22e+17 & 1.67e+19 & 5.34e+20 \\
			5 & 4.71e+10 & 6.70e+14 & 4.23e+16 & 6.53e+20 & 4.17e+22 & 2.67e+24 \\
			6 & 9.73e+11 & 6.77e+16 & 8.50e+18 & 6.54e+23 & 8.35e+25 & 1.07e+28 \\
			7 & 1.67e+13 & 5.71e+18 & 1.43e+21 & 5.46e+26 & 1.39e+29 & 3.56e+31 \\
			8 & 2.44e+14 & 4.13e+20 & 2.05e+23 & 3.90e+29 & 1.99e+32 & 1.02e+35 \\
			9 & 3.10e+15 & 2.62e+22 & 2.59e+25 & 2.44e+32 & 2.49e+35 & 2.55e+38 \\
			10 & 3.46e+16 & 1.47e+24 & 2.90e+27 & 1.36e+35 & 2.77e+38 & 5.66e+41 \\
		\bottomrule
	\end{tabular}
	\label{fig:operations}
\end{table}

While the assumptions underlying \autoref{fig:operations} are idealized, it offers insights into the computational cost for CART-ELC to identify and split on a single decision node. These values indicate using CART-ELC on small datasets ($n \leq 1000$) when $r$ values are small ($r \leq 3$) remains manageable, but as $r$ and $n$ values increase, the computational costs quickly grow reaching $3.13 \cdot 10^{14}$ operations at $r = 3$, $n = 5000$ for a single split. 

While performing our evaluations shown in \autoref{fig:results}, we found $r$ values of three or higher to be too costly in most cases. This was partially because the results shown in \autoref{fig:operations} assume $r=m$, but in \autoref{fig:results} this was often not true, and thus there was an additional multiplicative factor of $\binom{m}{r}$.

\section{Experiments and Results}

In this section, we present our data collection methodology and provide a detailed description of the datasets used. We then compare CART-ELC with other decision tree induction algorithms, evaluate different splitting criteria, and discuss our findings.

\subsection{Methodology}\label{methodology}

The results shown in \autoref{fig:results} for OC1 and OC1-AP \citep{MurthyKS94}, CART-LC and CART \citep{breiman1984cart}, and C4.5 \citep{c4.5}, where C4.5 and CART were collected using the IND 2.1 package \citep{buntine}, were reported by \citet{MurthyKS94}. Alongside these, we present results from our implementations of HHCART \citep{HHCart} and CART-ELC.%
\makeatletter
\@ifpackagewith{tmlr}{accepted}{\footnote{Our implementations of HHCART and CART-ELC are available at \url{https://github.com/andrewlaack/cart-elc}.}}{}%
\@ifpackagewith{tmlr}{preprint}{\footnote{Our implementations of HHCART and CART-ELC are available at \url{https://github.com/andrewlaack/cart-elc}.}}{}%
\makeatother
\space To ensure a fair comparison, we followed experimental procedures consistent with those of \citet{MurthyKS94}, differing only in our hyperparameter tuning strategy for CART-ELC and HHCART. The change to hyperparameter tuning was necessary due to our inclusion of a max depth hyperparameter for HHCART and CART-ELC, as well as the $r$ hyperparameter for CART-ELC.

Similar to the experiments performed by \citet{MurthyKS94}, HHCART and CART-ELC used the twoing criterion as the splitting criterion, defined in \autoref{splitting_criteria}. Additionally, HHCART and CART-ELC did not apply any post-pruning or pre-pruning strategies.
 
For each experiment with CART-ELC we defined two variables, $d$ and $j$, the largest max depth and $r$ value to test respectively. We set $d = 5$ because increasing the max depth beyond five often led to identical tree structures or overfitting (see \autoref{hyper_tuning}). We also set $j = 2$ for all datasets except for the iris dataset where we set $j = 4$. These choices for $j$ values were made due to computational constraints imposed by our algorithm's complexity and dataset sizes, outlined in \autoref{complexity}. 

Shown below are the steps performed for each experiment.

\begin{enumerate}
	\item Complete ten iterations of the following:
    \begin{enumerate}
        \item Randomly partition the dataset into five partitions of roughly equal size.
		\item Perform a grid search over all integers $1 \leq i \leq j$ and $1 \leq k \leq d$ where $k$ is defined as the current max depth hyperparameter, and $i$ is defined as the current $r$ hyperparameter. For each iteration:
		\begin{enumerate}
			\item Build a tree using all partitions except for one as the training data.
			\item Evaluate the accuracy of the tree on the unseen partition.
			\item Repeat steps (i) and (ii) for each of the five partitions.
			\item Calculate the mean accuracy and tree size across the five iterations where tree size is defined as the number of leaf nodes in the tree.
            \end{enumerate}
    \end{enumerate}
\item Calculate the mean and standard deviation of the tree sizes and accuracies based on the final results of each cross-validation (CV) \citep{cv} iteration for each hyperparameter pair.
    \item Select the hyperparameter pair that offers the best balance between mean accuracy and tree size. The mean and standard deviation of the accuracy and tree size with the selected hyperparameter pair are the results of the experiment.
\end{enumerate}

For our evaluation of both HHCART algorithms, we performed the same steps outlined above, but HHCART does not have an $r$ hyperparameter, and thus our hyperparameter search was only performed for max depth values.

\subsection{Datasets}\label{datasets}

\subsubsection{Star/Galaxy Discrimination (Bright)}\label{sg_bright}

This dataset was derived from bright images collected by \citet{odewahn} of stars and galaxies. The dataset is comprised of 2,462 samples, 14 features, and two classifications. The two classifications are star and galaxy. The 14 features were measurements defined by the original researchers that they deemed to be relevant for star/galaxy discrimination.

\subsubsection{Star/Galaxy Discrimination (Dim)}\label{sg_dim}

This dataset was derived from dim images collected by \citet{odewahn} of stars and galaxies. The dataset is comprised of 4,192 samples, 14 features, and two classifications. The two classifications are star and galaxy. The 14 features were measurements defined by the original researchers that they deemed to be relevant for star/galaxy discrimination.

\subsubsection{Breast Cancer Diagnosis}\label{Breast Cancer Diagnosis}

This dataset \citep{breast_cancer} is comprised of 699 samples with nine features and two classifications. The two classifications are malignant and benign. Similar to the evaluations performed by \citet{MurthyKS94}, we removed the 16 samples from this dataset that were missing the "Bare Nuclei" feature. Of the remaining 683 samples, 444 were labeled as benign and 239 were labeled as malignant.

\subsubsection{Iris Classification}\label{iris}

This dataset \citep{iris} is comprised of 150 samples with four features and three classifications. Each of the three classifications specifies a different species of iris and contains 50 samples.

\subsubsection{Boston Housing Cost}\label{housing}

This dataset \citep{boston} is comprised of 506 samples with 12 continuous features, one binary feature, and two classifications. While the original dataset contained a continuous label, we discretized the dataset by assigning the class of a given sample to be one if the continuous label's value was strictly less than 21,000 and two if it was greater than or equal to 21,000, as was done by \citet{MurthyKS94}. After this preprocessing, the dataset contained 260 samples with a classification of two and 246 samples with a classification of one. The dataset was also missing some features for select samples which we left as NaN for our evaluation of CART-ELC and performed mean imputation for HHCART as HHCART is unable to handle missing features.

\subsubsection{Diabetes Diagnosis}\label{diabetes}

This dataset \citep{diabetes} is comprised of 768 samples with eight features and two classifications. These two classifications represent testing positive and negative for diabetes. Of the 768 samples, 500 tested negative for diabetes, and 268 tested positive.

\newpage
\subsection{Empirical Algorithmic Comparison}\label{empirical_comparison}

\begin{table}[h]
	\centering
	\small
    \caption{Accuracy and tree size comparison across decision tree induction algorithms.}
	\begin{tabular}{lcccccc} 
		\addlinespace
		\toprule
		\multirow{2}{*}{Algorithm} & \multicolumn{6}{c}{Accuracy}  \\ 
		\cmidrule(lr){2-7}  
		& S/G Bright & S/G Dim & Cancer & Iris & Housing & Diabetes \\ 
		\midrule
		CART-ELC  & \textbf{98.9 ± 0.2} & \textbf{95.2 ± 0.5} & 96.3 ± 0.4 & 95.1 ± 0.8 & 83.5 ± 0.7 & \textbf{74.5 ± 1.3}  \\

		HHCART(A) & 98.3 ± 0.5 & 93.7 ± 0.8 & \textbf{96.9 ± 0.3} &  \textbf{95.5 ± 1.4} &  \textbf{83.9 ± 0.8} & 73.2 ± 1.2  \\
		HHCART(D) & 98.1 ± 0.4 & 93.7 ± 0.9 & \textbf{96.9 ± 0.3} &  94.3 ± 1.5 &  82.2 ± 1.4 & 73.2 ± 1.2  \\
		OC1       & \textbf{98.9 ± 0.2} & 95.0 ± 0.3 & 96.2 ± 0.3 & 94.7 ± 3.1 & 82.4 ± 0.8 & 74.4 ± 1.0  \\
		OC1-AP    & 98.1 ± 0.2 & 94.0 ± 0.2 & 94.5 ± 0.5 & 92.7 ± 2.4 & 81.8 ± 1.0 & 73.8 ± 1.0  \\
		CART-LC   & 98.8 ± 0.2 & 92.8 ± 0.5 & 95.3 ± 0.6 & 93.5 ± 2.9 & 81.4 ± 1.2 & 73.7 ± 1.2  \\
		CART   & 98.5 ± 0.5 & 94.2 ± 0.7 & 95.0 ± 1.6 & 93.8 ± 3.7 & 82.1 ± 3.5 & 73.9 ± 3.4  \\
		C4.5      & 98.5 ± 0.5 & 93.3 ± 0.8 & 95.3 ± 2.0 & 95.1 ± 3.2 & 83.2 ± 3.1 & 71.4 ± 3.3  \\ 
		\midrule
		\multirow{2}{*}{Algorithm} & \multicolumn{6}{c}{Tree Size}  \\ 
		\cmidrule(lr){2-7}  
		& S/G Bright & S/G Dim & Cancer & Iris & Housing & Diabetes \\ 
		\midrule
		CART-ELC  & \textbf{3.7 ± 0.2}  & \textbf{9.8 ± 4.2}  & \textbf{2.0 ± 0.0}  & 4.8 ± 0.1  & \textbf{4.0 ± 0.0}  & \textbf{4.0 ± 0.0}  \\
		HHCART(A) & 6.1 ± 0.3  & 14.6 ± 4.8  & \textbf{2.0 ± 0.0}  & \textbf{3.1 ± 0.1}  & 7.8 ± 0.2  & \textbf{4.0 ± 0.0}  \\
		HHCART(D) & 6.3 ± 0.4  & 14.9 ± 5.0  & \textbf{2.0 ± 0.0}  & 4.7 ± 0.1  & 23.3 ± 0.8  & \textbf{4.0 ± 0.0}  \\
		OC1       & 4.3 ± 1.0  & 13.0 ± 8.7  & 2.8 ± 0.9  & \textbf{3.1 ± 0.2}  & 6.9 ± 3.2  & 5.4 ± 3.8  \\
		OC1-AP    & 6.9 ± 2.4  & 29.3 ± 8.8  & 6.4 ± 1.7  & 3.2 ± 0.3  & 8.6 ± 4.5  & 11.4 ± 7.5  \\
		CART-LC   & 3.9 ± 1.3  & 24.2 ± 8.7  & 3.5 ± 0.9  & 3.2 ± 0.3  & 5.8 ± 3.2  & 8.0 ± 5.2  \\
		CART   & 13.9 ± 5.7  & 30.4 ± 10.0  & 11.5 ± 7.2  & 4.3 ± 1.6  & 15.1 ± 10  & 11.5 ± 9.1  \\
		C4.5      & 14.3 ± 2.2  & 77.9 ± 7.4  & 9.8 ± 2.2  & 4.6 ± 0.8  & 28.2 ± 3.3  & 56.3 ± 7.9  \\ 
		\bottomrule
	\end{tabular}
	\label{fig:results}
\end{table}

The results presented in \autoref{fig:results} demonstrate that CART-ELC, HHCART(A), HHCART(D), and OC1 obtained consistently strong accuracies across the datasets evaluated when compared with other decision tree induction algorithms. Additionally, of these top performers, CART-ELC had the smallest trees across most datasets, falling behind on the iris dataset where a higher max depth value was selected due to minor variances in accuracies across max depth and $r$ values. Finally, we note the standard deviations in accuracy for CART-ELC, on average, were lower than those of OC1 and the variants of HHCART.

\begin{table}[h]
	\centering
\caption{P-values for accuracy comparisons between various models and CART-ELC. P-values strictly less than 0.05 are bolded. P-values reported as 0.000 are not zero, but are smaller than the reporting precision.}
	\small
	\begin{tabular}{lccccccc} 
		\addlinespace
		\toprule
		\multirow{1}{*}{Algorithm} & \multirow{1}{*}{S/G Bright} & \multirow{1}{*}{S/G Dim} & \multirow{1}{*}{Cancer} & \multirow{1}{*}{Iris} & \multirow{1}{*}{Housing} & \multirow{1}{*}{Diabetes} \\ 
		\midrule
HHCART(A) & \textbf{0.004} & \textbf{0.000} & \textbf{0.001} & 0.446 & 0.250 & \textbf{0.032} \\
HHCART(D) & \textbf{0.000} & \textbf{0.000} & \textbf{0.001} & 0.159 & \textbf{0.021} & \textbf{0.032} \\
OC1 & 1.000 & 0.296 & 0.536 & 0.701 & \textbf{0.004} & 0.849 \\
OC1-AP & \textbf{0.000} & \textbf{0.000} & \textbf{0.000} & \textbf{0.012} & \textbf{0.000} & 0.195 \\
CART-LC & 0.278 & \textbf{0.000} & \textbf{0.000} & 0.122 & \textbf{0.000} & 0.170 \\
CART & \textbf{0.037} & \textbf{0.002} & \textbf{0.032} & 0.303 & 0.244 & 0.612 \\
C4.5 & \textbf{0.037} & \textbf{0.000} & 0.153 & 1.000 & 0.771 & \textbf{0.017} \\
    \bottomrule
\end{tabular}
\label{fig:p_values}
\end{table}

The results in \autoref{fig:p_values}, obtained by performing Welch's t-tests, compare the accuracies of each algorithm against CART-ELC. These results support the finding that CART-ELC, HHCART, and OC1 are the best-performing decision tree induction algorithms on the datasets evaluated. Additionally, CART-ELC had statistically significant performance improvements on both star/galaxy discrimination datasets and the diabetes dataset relative to both HHCART variants and only had statistically significant performance improvements when compared to OC1 on the housing dataset. The final result of interest is that both variants of HHCART performed better than CART-ELC on the cancer dataset by a statistically significant amount.

\begin{table}[h]
	\centering
	\caption{Cohen's d effect size for accuracies between various models and CART-ELC. Comparisons with $p < 0.05$ are bolded.}
		\small
	\begin{tabular}{lccccccc} 
		\addlinespace
		\toprule
		\multirow{1}{*}{Algorithm} & \multirow{1}{*}{S/G Bright} & \multirow{1}{*}{S/G Dim} & \multirow{1}{*}{Cancer} & \multirow{1}{*}{Iris} & \multirow{1}{*}{Housing} & \multirow{1}{*}{Diabetes} \\ 
		\midrule

		HHCART(A) & \textbf{1.576} & \textbf{2.249} & \textbf{-1.697} & -0.351 & -0.532 & \textbf{1.039} \\
		HHCART(D) & \textbf{2.530} & \textbf{2.060} & \textbf{-1.697} & 0.666 & \textbf{1.175} & \textbf{1.039} \\
	OC1 & 0.000 & 0.485 & 0.283 & 0.177 & \textbf{1.463} & 0.086 \\
	OC1-AP & \textbf{4.000} & \textbf{3.151} & \textbf{3.976} & \textbf{1.342} & \textbf{1.970} & 0.604 \\
	CART-LC & 0.500 & \textbf{4.800} & \textbf{1.961} & 0.752 & \textbf{2.138} & 0.639 \\
	CART & \textbf{1.050} & \textbf{1.644} & \textbf{1.115} & 0.486 & 0.555 & 0.233 \\
	C4.5 & \textbf{1.050} & \textbf{2.848} & 0.693 & 0.000 & 0.133 & \textbf{1.236} \\
    \bottomrule
\end{tabular}
\label{fig:cohens_d}
\end{table}

The results in \autoref{fig:cohens_d} show the effect sizes of the difference between accuracies with respect to CART-ELC. Negative values indicate the algorithm outperformed CART-ELC on the dataset, and positive values indicate CART-ELC outperformed the algorithm on the dataset. These results show the magnitude of CART-ELC's outperformance of HHCART on the star/galaxy datasets is greater than the magnitude of HHCART's outperformance of CART-ELC on the cancer dataset. 

HHCART's strong performance relative to CART-ELC on the cancer dataset is likely due to the dataset being both linearly separable and having a sizable gap between the two classes. HHCART typically uses splits that pass through one sample in the transformed feature space, and thus it can find splits with a larger separation on both sides with respect to the classes on each side. This contrasts CART-ELC where each hyperplane generally passes through multiple samples and will thus generalize worse on some datasets because the separation on both sides of the hyperplane will be smaller.

Despite this, a limitation of HHCART that makes CART-ELC perform better on certain datasets is its inability to handle outliers well. Outliers nominally impact CART-ELC as the splitting criterion is minimally impacted by them, but HHCART uses a change of basis derived from the eigen decomposition of the covariance matrix for each class, and these resulting eigenvectors can be greatly impacted by outliers. This results in HHCART searching for axis-aligned splits in suboptimal feature spaces. Similarly, when there are clusters of samples with the same classification throughout the feature space, this can also pose issues for HHCART as the covariance matrices will be impacted by the disparate groups more than CART-ELC will be.

\subsection{Empirical Criteria Comparison}\label{empirical_criteria_comparison}

\begin{table}[h]
	\centering
    \caption{Accuracy and tree size comparison across splitting criteria.}
	\small
	\begin{tabular}{lccccccc} 
		\addlinespace
		\toprule
		\multirow{2}{*}{Splitting Criterion} & \multicolumn{6}{c}{Accuracy}  \\ 
		\cmidrule(lr){2-7}  
		\multirow{1}{*}{} & \multirow{1}{*}{S/G Bright} & \multirow{1}{*}{S/G Dim} & \multirow{1}{*}{Cancer} & \multirow{1}{*}{Iris} & \multirow{1}{*}{Housing} & \multirow{1}{*}{Diabetes} \\ 
		\midrule
		\hyperref[twoing]{Twoing Criterion} & \textbf{98.9 ± 0.2} & \textbf{95.2 ± 0.5} & \textbf{96.3 ± 0.4} & 95.1 ± 0.8 & \textbf{83.5 ± 0.7} & \textbf{74.5 ± 1.3} \\

		\hyperref[gini]{Gini Criterion} & \textbf{98.9 ± 0.3} & 95.1 ± 0.5 & \textbf{96.3 ± 0.4} & 95.1 ± 0.9 & \textbf{83.5 ± 0.7} & \textbf{74.5 ± 1.3} \\

		\hyperref[IG]{Information Gain}& \textbf{98.9 ± 0.2} & 94.9 ± 0.5 & 96.0 ± 0.4 & \textbf{95.3 ± 0.8} & 81.8 ± 1.0 & 74.0 ± 1.1 \\

		\midrule
		\multirow{2}{*}{Splitting Criterion} & \multicolumn{6}{c}{Tree Size}  \\ 
		\cmidrule(lr){2-7}  
		& S/G Bright & S/G Dim & Cancer & Iris & Housing & Diabetes \\ 
		\midrule

		\hyperref[twoing] {Twoing Criterion} & 3.7 ± 0.2 & \textbf{9.8 ± 4.2} & \textbf{2.0 ± 0.0} & 4.8 ± 0.1 & 4.0 ± 0.0 & \textbf{4.0 ± 0.0} \\ 
			\hyperref[gini]{Gini Criterion}	& \textbf{3.6 ± 0.1} & 12.4 ± 8.3 & \textbf{2.0 ± 0.0} & \textbf{3.0 ± 0.1} & 4.0 ± 0.0 & \textbf{4.0 ± 0.0} \\
		\hyperref[IG]{Information Gain} & 3.8 ± 0.2 & 9.9 ± 4.5 & \textbf{2.0 ± 0.0} & 4.8 ± 0.1 & \textbf{2.0 ± 0.0} & \textbf{4.0 ± 0.0} \\ 

    \bottomrule
	\end{tabular}
	\label{fig:crits}
\end{table}

Our experimental results across splitting criteria, shown in \autoref{fig:crits}, indicate the twoing criterion and Gini criterion are both strong splitting criteria, and information gain only falls behind nominally. Additionally, on the dataset we evaluated that contained more than two classifications, information gain outperformed the other two criteria.

\section{Conclusions and Future Work}\label{conclusions_and_future_work}

CART-ELC achieves competitive accuracies across a variety of small datasets when compared with other decision tree algorithms despite only searching a subset of all unique hyperplanes. Additionally, while other oblique decision trees often use splits that are linear combinations of many features, CART-ELC's $r$ hyperparameter limits the number of features in each linear combination, which can result in simpler decision boundaries. This, along with CART-ELC's ability to generate shallower trees, often results in trees that are both more interpretable and efficient when making predictions.

Despite this, the computational complexity of CART-ELC remains prohibitively high for large datasets. As such, future research should focus on strategies to mitigate these computational costs, such as developing methods to select a smaller subset of candidate hyperplanes or using CART-ELC with ensemble methods. Adapting CART-ELC for use in random forests \citep{rndfrst} and stochastic gradient boosting \citep{sgboost} may be a promising direction, as these approaches could reduce computational costs by training on subsets of samples or by introducing stochastic elements into the training process. Finally, while our evaluation included six datasets, further experimentation on a wider variety of datasets and more extensive CV may be beneficial for assessing the algorithm's overall efficacy.

\bibliography{references}
\bibliographystyle{tmlr}

\appendix

\newpage

\section{Hyperparameter Tuning Results for CART-ELC}\label{hyper_tuning}

\begin{figure}[h]
\includegraphics[width=\linewidth]{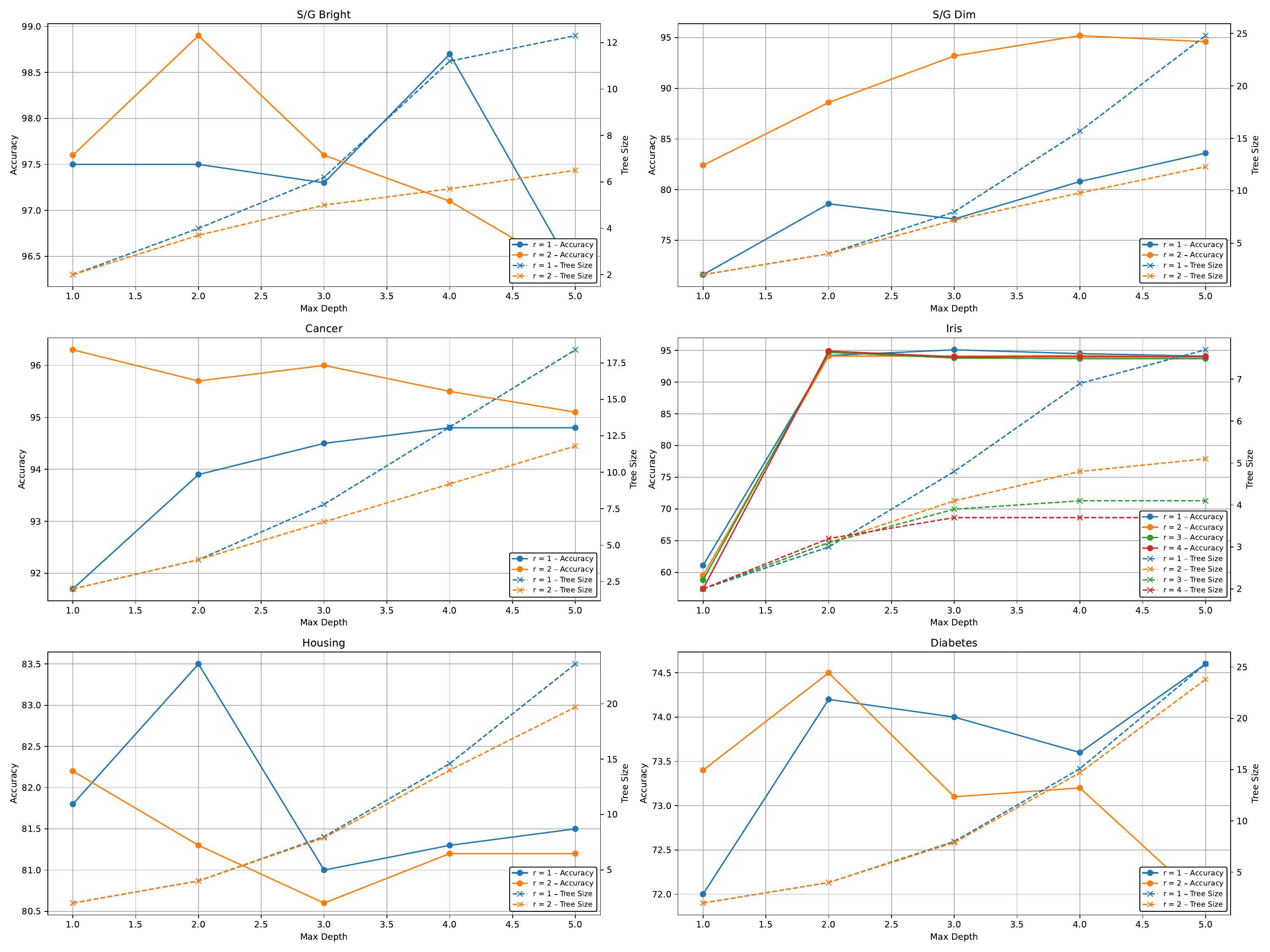}
\caption{Max depth and $r$ value results for CART-ELC when using the twoing criterion as the splitting criterion.}
\label{graph_twoing}
\end{figure}

\begin{figure}[h]
	\includegraphics[width=\linewidth]{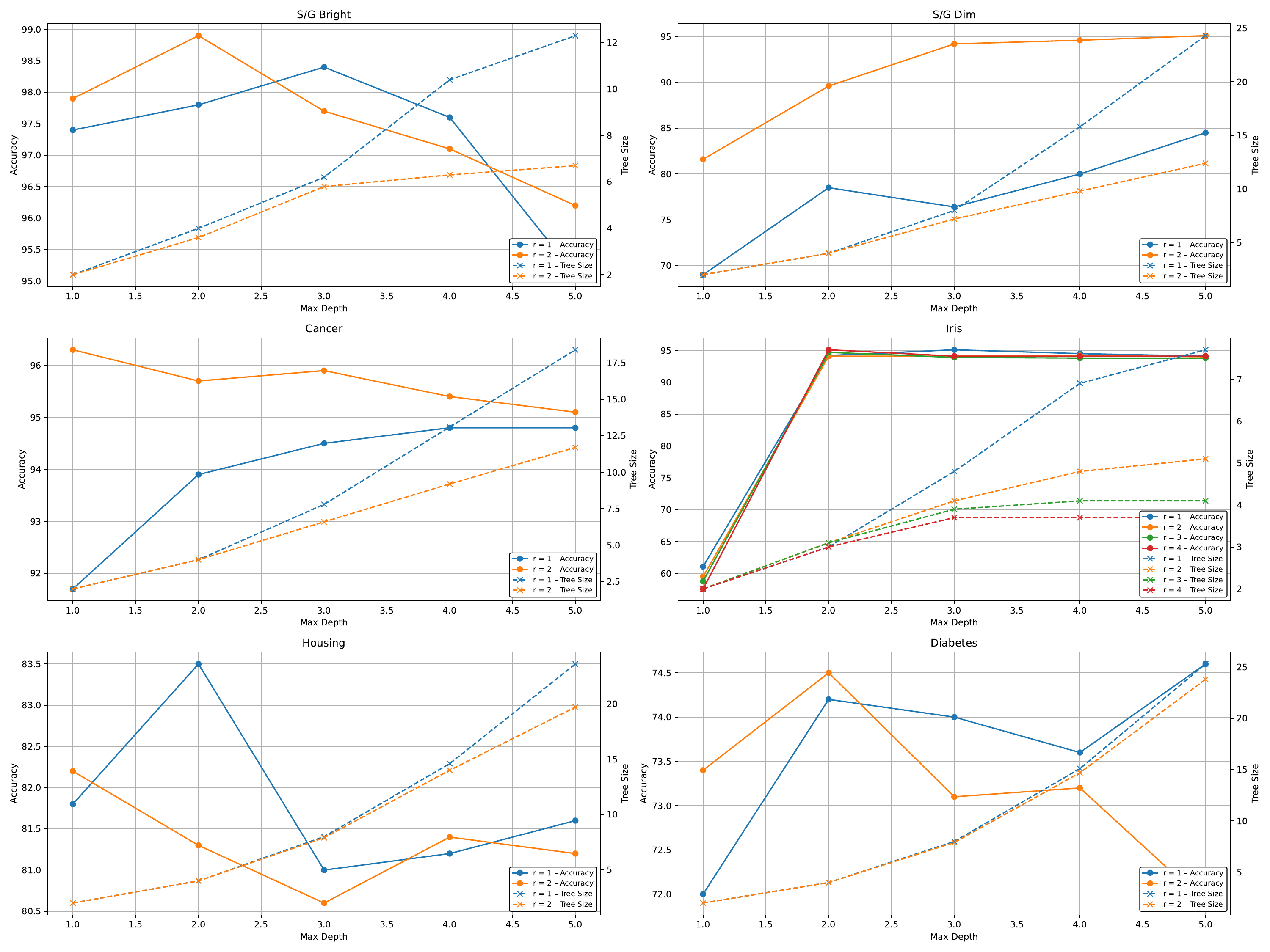}
	\caption{Max depth and $r$ value results for CART-ELC when using the Gini criterion as the splitting criterion.}
\label{graph_gini}
\end{figure}

\begin{figure}[h]
\includegraphics[width=\linewidth]{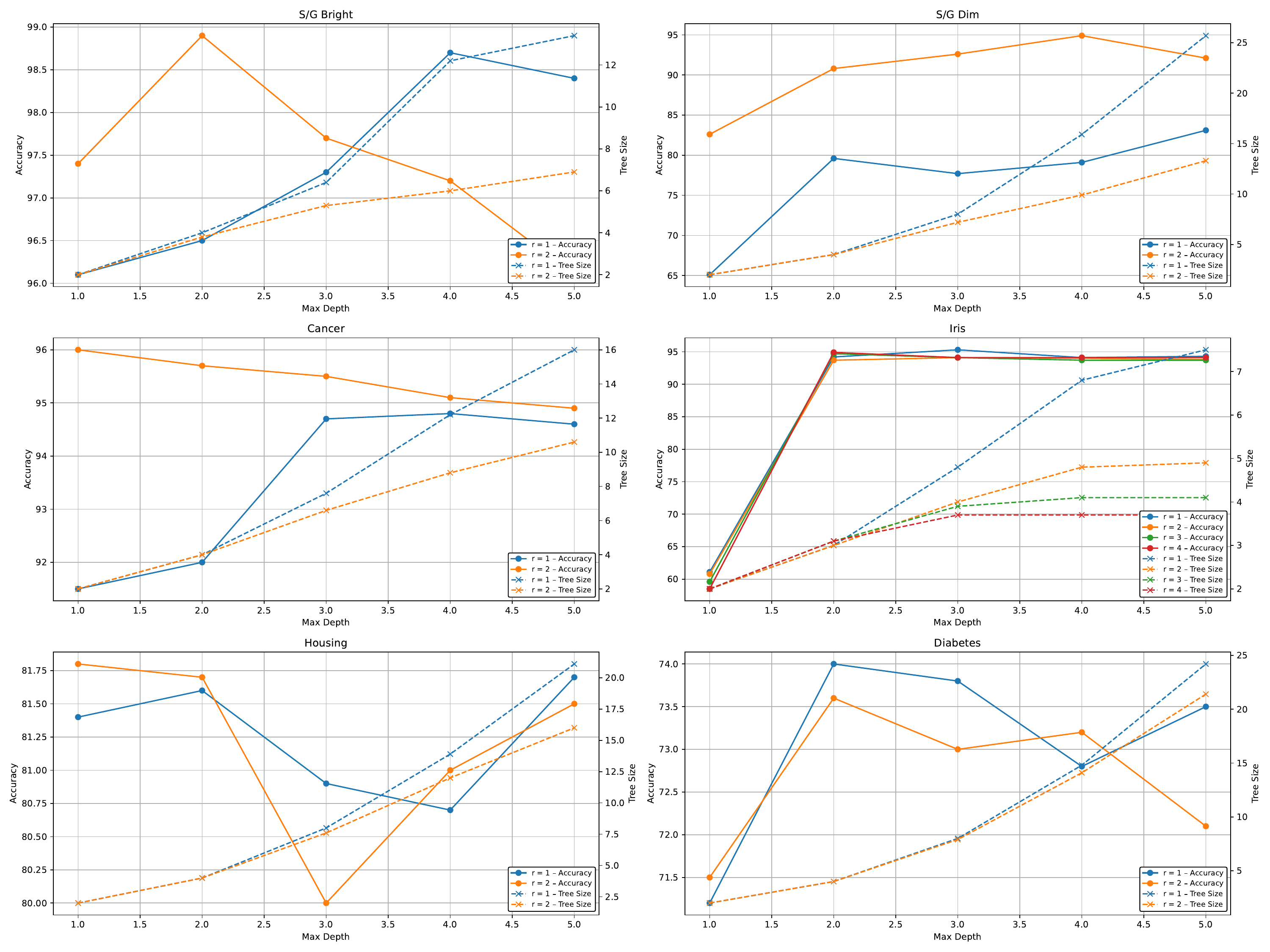}
	\caption{Max depth and $r$ value results for CART-ELC when using information gain as the splitting criterion.}
\label{graph_ig}
\end{figure}

Figures~\ref{graph_twoing}--\ref{graph_ig} show the accuracies and tree sizes for CART-ELC across each splitting criterion, dataset, $r$ value, and max depth. These results show higher $r$ values often result in higher accuracies at lower max depths but begin overfitting at lower max depths. We also see the accuracies and tree sizes across splitting criteria are quite consistent, often varying in accuracy but rarely in trends. Finally, on the S/G Dim dataset, accuracies appear to be plateauing around a max depth of five for $r=2$ while on the other datasets, max depths of five either result in models overfitting or accuracies stagnating for $r \geq 2$.

\end{document}